\documentclass[sigconf]{acmart}
\usepackage{multirow}
\usepackage{colortbl}
\usepackage{multirow}
\usepackage{algorithm}
\usepackage{algorithmicx}
\usepackage{algpseudocode}

\floatname{algorithm}{Algorithm}

\usepackage[normalem]{ulem}
\useunder{\uline}{\ul}{}

\AtBeginDocument{%
  }

\setcopyright{rightsretained}
\copyrightyear{2024}
\acmYear{2024}
\acmDOI{10.1145/3664647.3680673}
\acmConference[MM '24] {Proceedings of the 32nd ACM International Conference on Multimedia}{October 28--November 1, 2024}{Melbourne, VIC, Australia.}
\acmBooktitle{Proceedings of the 32nd ACM International Conference on Multimedia (MM '24), October 28--November 1, 2024, Melbourne, VIC, Australia}

\acmISBN{979-8-4007-0686-8/24/10}

\makeatletter
\gdef\@copyrightpermission{
 \begin{minipage}{0.3\columnwidth}
 \href{https://creativecommons.org/licenses/by/4.0/}{\includegraphics[width=0.90\textwidth]{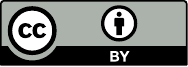}}
 \end{minipage}\hfill
 \begin{minipage}{0.7\columnwidth}
 \href{https://creativecommons.org/licenses/by/4.0/}{This work is licensed under a Creative Commons 
Attribution International 4.0 License.}
 \end{minipage}
 \vspace{5pt}
}
\makeatother
\begin{document}

\title{T2VIndexer: A Generative Video Indexer for Efficient Text-Video Retrieval}


\author{Yili Li}
\affiliation{%
  \institution{Institute of Information Engineering, Chinese Academy of Sciences}
  \institution{School of Cyber Security, University of Chinese Academy of Sciences}
  \city{Beijing}
  \country{China}}
\email{liyili@iie.ac.cn}
\orcid{0009-0006-4037-819X}


\author{Jing Yu}
\authornotemark[2]
\affiliation{%
  \institution{Institute of Information Engineering, Chinese Academy of Sciences}
  \institution{School of Cyber Security, University of Chinese Academy of Sciences}
  \city{Beijing}
  \country{China}}
\email{yujing02@iie.ac.cn}
\orcid{0000-0002-3966-511X}

\author{Keke Gai}
\affiliation{%
  \institution{School of Cyberspace Science and Technology, Beijing Institute of Technology}
  \city{Beijing}
  \country{China}}
\email{gaikeke@bit.edu.cn}
\orcid{0000-0001-6784-0221}

\author{Bang Liu}
\affiliation{%
  \institution{Université de Montréal \& Mila}
  \city{Montréal}
  \country{Canada}}
\email{bang.liu@umontreal.ca}
\orcid{0000-0002-9483-8984}

\author{Gang Xiong}
\affiliation{%
  \institution{Institute of Information Engineering, Chinese Academy of Sciences}
  \city{Beijing}
  \country{China}}
\email{xionggang@iie.ac.cn}
\orcid{0000-0002-3190-6521}

\author{Qi Wu}
\affiliation{%
  \institution{Australia Institute of Machine Learning, University of Adelaide}
  \city{Adelaide}
  \country{Australia}}
\email{qi.wu01@adelaide.edu.au}
\orcid{0000-0003-3631-256X}
\renewcommand{\shortauthors}{Yili Li and Jing Yu, et al.}

\begin{abstract}
  Current text-video retrieval methods mainly rely on cross-modal matching between queries and videos to calculate their similarity scores, which are then sorted to obtain retrieval results. This method considers the matching between each candidate video and the query, but it incurs a significant time cost and will increase notably with the increase of candidates. Generative models are common in natural language processing and computer vision, and have been successfully applied in document retrieval, but their application in multimodal retrieval remains unexplored. To enhance retrieval efficiency, in this paper, we introduce a model-based video indexer named T2VIndexer, which is a sequence-to-sequence generative model directly generating video identifiers and retrieving candidate videos with constant time complexity. T2VIndexer aims to reduce retrieval time while maintaining high accuracy. To achieve this goal, we propose video identifier encoding and query-identifier augmentation approaches to represent videos as short sequences while preserving their semantic information. Our method consistently enhances the retrieval efficiency of current state-of-the-art models on four standard datasets. It enables baselines with only 30\%-50\% of the original retrieval time to achieve better retrieval performance on MSR-VTT (+1.0\%), MSVD (+1.8\%), ActivityNet (+1.5\%), and DiDeMo (+0.2\%). The code is available at \href{https://github.com/Lilidamowang/T2VIndexer-generativeSearch}{https://github.com/Lilidamowang/T2VIndexer-generativeSearch}.
\end{abstract}

\ccsdesc[500]{Information systems~Language models}
\ccsdesc[500]{Information systems~Retrieval models and ranking}
\ccsdesc[500]{Information systems~Novelty in information retrieval}
\keywords{Deep Learning, Multi-modal Learning, Video Retrieval, Generative Model}   



\maketitle

\section{Introduction}
\label{sec:intro}
Given a query text description, text-video retrieval \cite{xu2016msr} aims to retrieve videos that are semantically relevant to the query. Text-video retrieval is flexible to express the user's intent and brings emerging attention for web search with the dramatic increasing of videos uploade d online every day. For a standard web search engine \cite{DBLP:conf/acl/RenZLWWW23}, video retrieval and ranking are two core stages. The retrieval stage first retrieves limited number of candidate videos from massive online videos, and the following ranking stage predicts accurate ranking scores between per query and the candidate videos. Since videos have much richer and more diverse visual content compared with the query text, precise video ranking is costly for fine-grained text-video matching. Therefore, the efficiency and recall performance of the video retrieval stage are crucial for achieving fast and accurate text-video search results.

\begin{figure}[t]
  \centering
  \includegraphics[width=\linewidth]{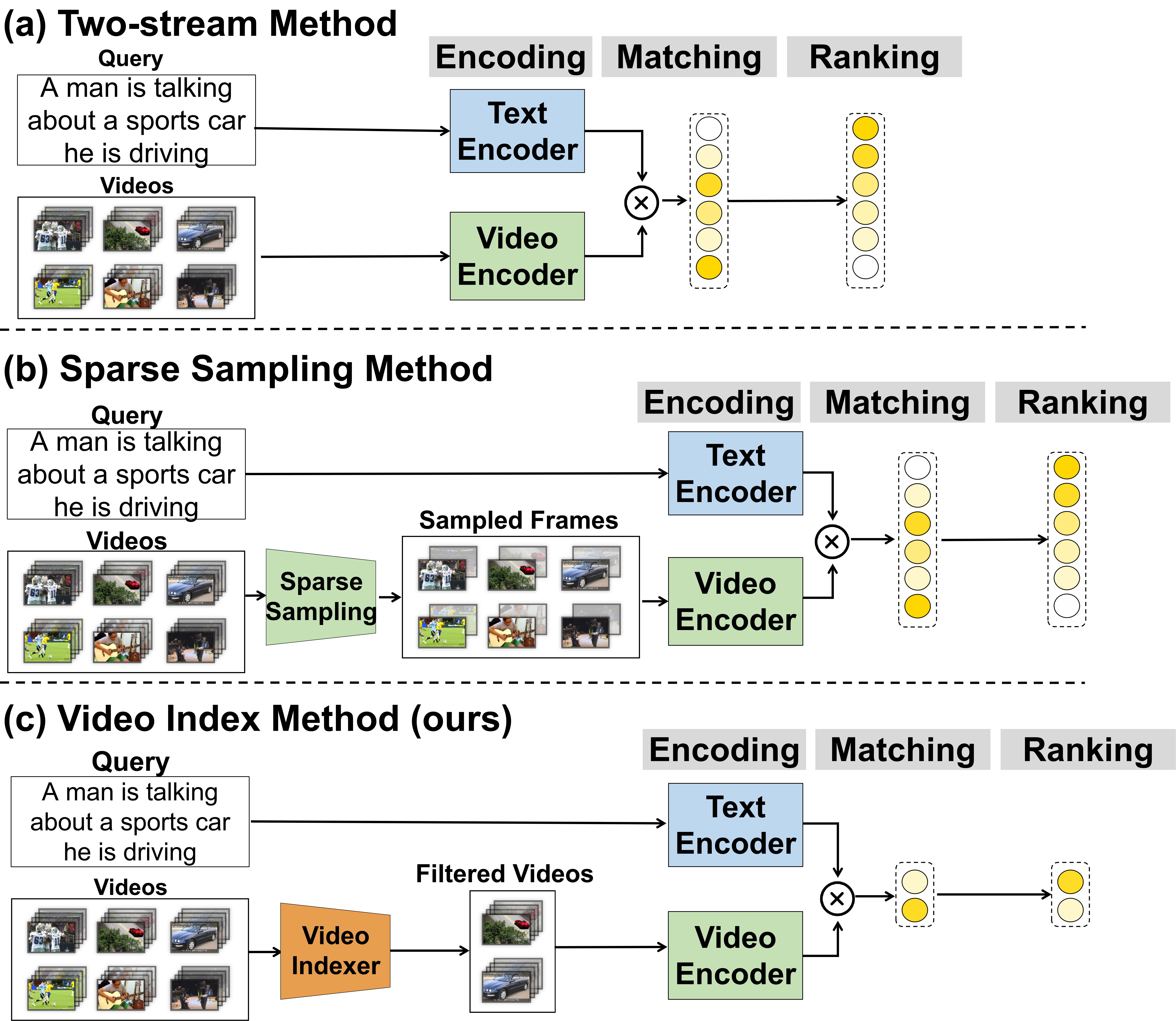}
  \caption{(a) Two stream method with independent video and text encoders. (b) Video sparse sampling for efficiency boost. (c) Our generative video indexer for efficiency boost.}
  \label{fig:mov}
  \vspace{-10pt}
\end{figure}

Existing text-video retrieval methods can be divided into two categories, namely \textit{one-stream} and \textit{two-stream} approaches. One-stream approaches \cite{zhu2020actbert} \cite{lei2021less} \cite{luo2020univl} adopts deep models for feature-level  interactions between each text-video pair to predict its similarity score, which require online feature extraction and fail to be applied for the time-sensitive retrieval stage. Thus, the efficient two-stream approaches \cite{DBLP:conf/eccv/Gabeur0AS20} \cite{DBLP:journals/ijon/LuoJZCLDL22} \cite{DBLP:conf/cvpr/WuLFWO23} are widely applied. As shown in Figure \ref{fig:mov} (a), they encode each video and text independently into dense embeddings and then adopt simple matching functions to measure their similarity. Since there are no text-video interactions in the encoding stage, two-stream approaches allows offline data embedding extraction and alleviating online computation. Some recent works begin to focus on the issues of reducing the high computational overload of dense video embedding by sparsely sampling a few clips \cite{lei2021less} (see Figure \ref{fig:mov} (b)). However, all the existing solutions require to measure the query-video similarities and then rank videos for the entire video set (\textit{i.e.}, one-to-all retrieval framework). Thus, their online retrieval time grows linearly with the increase of retrieved videos, which limits their scalability on large-scale scenarios.  

To address the above issue, we explore to fundamentally change the traditional one-to-all embedding retrieval framework by a generative deep model that directly generates video identifiers and retrieves video candidates with constant time complexity. As illustrated in Figure \ref{fig:mov} (c), our target of this work is not to propose a new model on text-video retrieval. We mainly investigate how to design a model-based indexer that effectively retrieves query-relevant video candidates, which shortens the overall retrieval time while maintaining the retrieval accuracy of state-of-the-art ranking models. To this end, we propose a sequence-to-sequence generative network that supports \textbf{T}ext query to \textbf{V}ideo candidate \textbf{Index}, named as \textbf{T2VIndexer}. The model is based on the encoder-decoder that feeds the query into the encoder and generates the identifier of the video candidate through the decoder. It is trained by query-identifier pairs that supports controllable video recall at different semantic grained. During inference, the top $K$ videos are directly retrieved by beam search and identifier constrain. 

To guarantee the effectiveness of T2VIndexer, we have proposed several methods to tackle the key challenges. First, to get the semantic representations and coarse-to-fine identifiers of videos for controllable video recall, we utilize CLIP \cite{radford2021learning} to embed each video, and then cluster and encode the semantic embeddings in a hierarchical mode. Second, we propose to leverage the pre-trained multi-modal large language model \cite{DBLP:journals/corr/abs-2304-14178} to generate new queries with diverse views of the video content, which augments the query-identifier pairs during training for stronger generalization ability during inference. Third, we propose to train a generative network based on T5 \cite{DBLP:conf/nips/BrownMRSKDNSSAA20} architecture to enable the deep interactions between the query and video identifier, which enhances the cross-modal correlation learning for precise identifier prediction.

The main contributions are summarized as follows: 
(1) We propose a new sequence-to-sequence framework as a indexer for efficient video retrieval. Our method predicts candidate videos with constant time complexity, outperforming existing one-to-all embedding retrieval solutions with linear time complexity. It shows the effectiveness of generative video indexing and advances research on generation-based text-video retrieval mechanism.
(2) We propose the video identifier encoding and query-identifier augmentation approaches for learning T2VIndexer with strong generalization ability. T2VIndexer is model-agnostic and universal to cooperatewith various independent-embedding approaches.
(3) Our T2VIndexer method efficiently speeds up text-video retrieval, cutting time by 30\% to 50\% on standard tasks by cooperated with T2VIndex.

\vspace{-3pt}
\section{Related Work}
\label{sec:related}

\begin{figure*}[ht]

  \centering
  \includegraphics[width=\linewidth]{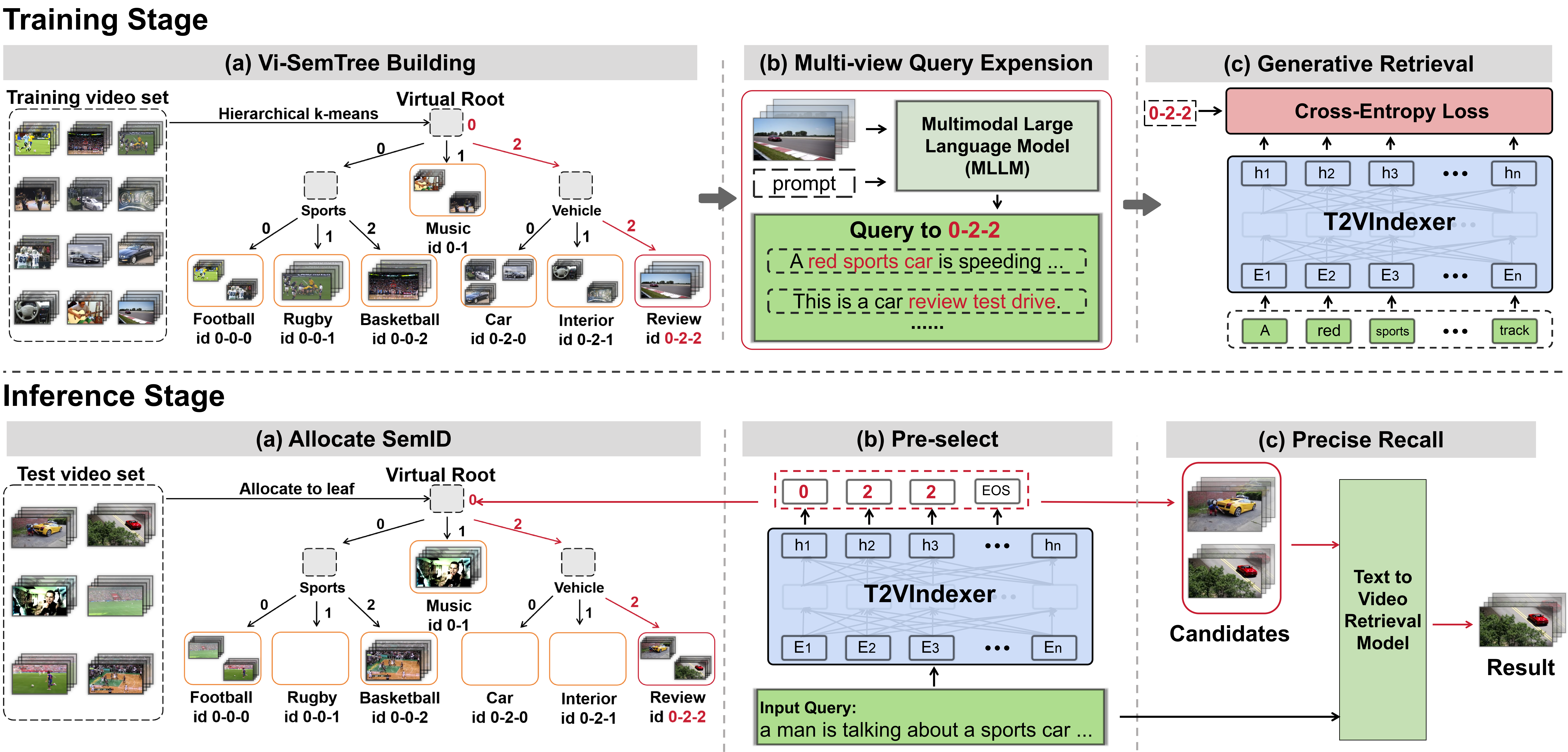}
  \caption{An overview of our T2VIndexer. T2VIndexer uses two different strategies for the training and inference stages. For the training stage, as shown in (a), the process of dividing the training set into a tree structure is illustrated. Training stage (b) shows the process of achieving multi-view query expansion through MLLM. (c) presents the pipeline for model training. For the inference stage, the new video is first inserted into the semantic tree and assigned a SemID, and the baseline model provides the precise retrieval results.}
  \label{fig:model}

\end{figure*}
\noindent \textbf{Text-video retrieval.}
Existing methods can be divided into two categories, called \textit{one-stream} and \textit{two-stream} approaches. One-stream approaches are characterized by token-level interactions based on cross-modal attention mechanisms, which are used for fine-grained video-text matching \cite{zhu2020actbert} \cite{lei2021less} \cite{luo2020univl} \cite{xu2021vlm}. Two-stream approaches aim to coordinate videos and text in a unified semantic space and perform direct comparisons through distance metrics \cite{DBLP:conf/eccv/Gabeur0AS20} \cite{DBLP:journals/ijon/LuoJZCLDL22} \cite{DBLP:conf/cvpr/WuLFWO23}. With the success of pre-trained image-text alignment model such as CLIP \cite{radford2021learning}, this method not only surpasses interactive embedding methods in efficiency but also has significant advantages in accuracy. In addition, efficiency enhancements have focused on video sampling strategies. Some methods choose to sample the frame sparsely \cite{lei2021less} \cite{DBLP:journals/ijon/LuoJZCLDL22}. Besides, redundancy persists within the vision tokens of each frame, diminishing the prowess of CLIP-style retrieval. CenterCLIP \cite{zhao2022centerclip} addressed this by refining patch subdivision and selection via clustering. These innovations enhance preprocessing efficiency but do not alleviate the inherent online retrieval latency due to similarity computations and ranking cost.

\noindent \textbf{Generative Model in Retrieval.}
In unimodal retrieval tasks, the same efficiency issues are faced. With the success of generative models in various visual and language tasks, they have demonstrated powerful capabilities. In document retrieval, models like DSI \cite{DBLP:conf/nips/Tay00NBM000GSCM22} demonstrate the ability to generate identifiers using Transformer architectures, while approaches like SEAL \cite{DBLP:conf/nips/BevilacquaOLY0P22} innovate by substituting string identifiers with document n-grams. The NCI \cite{DBLP:conf/nips/WangHWMWCXCZL0022} further refines this approach by integrating positional information into the decoding process. Image-to-image retrieval tasks have transformed these methods into visual modality. For example, IRGen \cite{DBLP:journals/corr/abs-2303-10126} tokenizes images to identifiers and uses a generative model to map queries to these identifiers for direct localization, thereby improving retrieval efficiency. 
In addition, \cite{DBLP:journals/corr/abs-2309-13375} proposed the SEATER retrieval framework, learning tree structured item identifiers via contrastive learning. Generative models were successfully applied in recommendation systems.
These methods have demonstrated powerful capabilities in unimodal retrieval. However, videos contain rich target and event. There is an obvious many-to-many problem, which means one video corresponds to multiple different descriptions from different perspectives, and a summary description corresponds to multiple different videos.

\section{Methodology}
\label{sec:methodology}

The text-video retrieval involves a text query $t$ and a gallery of videos $V$. The objective is to retrieve videos $\{v_j\} \in V$ that are semantically relevant to the query. As shown in Figure \ref{fig:model}, our goal is to directly retrieve targeted videos by generating the video identifiers based on natural language queries. To this end, we design a sequence-to-sequence generative model that takes query $t$ as input and outputs the video identifier for video index. We first propose a semantic-aware tree structure to encode video identifiers, called SemID, which encodes the multi-grained semantics of videos by a sequence for controllable recall while maintaining the sequence length as short as possible for fast encoding. To augment the semantic expression of queries for more generalized model learning, we propose to utilize a Multi-modal Large Language Model (MLLM) \cite{DBLP:journals/corr/abs-2304-14178} to generate a set of multi-view queries for each video, thereby enriching the contextual semantics encapsulated by the SemIDs for diverse queries. The model architecture, training and inference strategies are introduced in the end.

\subsection{Vi-SemTree for Video Identifying}
The purpose of our work is to locate videos by taking query $t$ as input and outputting the most relevant video identifier. Therefore, finding a suitable identifier as the basis for video location is crucial. The identifier needs to have semantic prior information so that it can reflect the content of the video, and similar semantic videos are also similar in the identifier. Moreover, the sequence length should be short enough to reduce the difficulty and complexity of model generation. Based on this consideration, we first extract the representation of each video, and construct a Video Semantic Tree (Vi-SemTree) based on the semantics of the video, and provide a SemID as an identifier for each video based on the tree structure. This approach ensures the semantic consistency of locating videos and guarantees the recall rate of the generation phase.

\noindent \textbf{Video Semantic Representation.}
To construct Vi-SemTree and obtain a sequence representation of SemID, we first extract the representation of the video. Compared with pixel-level information, Vi-SemTree requires the integration of semantic-level information, which is more seamlessly integrated with the structure of natural language. To meet this requirement, we chose the image encoder of CLIP \cite{radford2021learning}, which is famous for its multimodal pretraining ability, as the basic tool for our video encoding. Given a video’s sequence of frames $v_f=\{f^1, f^2, ..., f^N\}$, where $N$ is the number of frames. We derive the corresponding frame representation as $\hat{f}=\{\hat{f}^1, \hat{f}^2, ..., \hat{f}^N\}$, culminating in the overall video representation $\hat{F}$, obtained through mean pooling of the individual frame representations. Conforming to the methodologies laid out by ViT \cite{DBLP:conf/iclr/DosovitskiyB0WZ21} and CLIP \cite{radford2021learning}, the output gleaned from the [class] token is utilized to represent each frame.

\noindent \textbf{Hierarchical Vi-SemTree Building.}
We use a tree structure to encode videos, which helps preserve semantic information and ensures that similar semantic videos are also similar in the identifier. Moreover, by controlling the depth $d$ of the tree, we can achieve different levels of granularity and control the sequence length. Following NCI\cite{DBLP:conf/nips/WangHWMWCXCZL0022}, we use hierarchical $k$-means method for video feature $\hat{F}$, as shown in the training stage (a) of Figure \ref{fig:model}. First, we use the $k$-means algorithm to divide the training set videos into $k$ clusters based on their representation similarity. Each cluster is a tree node and contains a group of semantically similar videos, which serves as a layer of the tree structure. For each cluster, if the number of videos is greater than $c$, we use the $k$-means algorithm to further divide the cluster, generating the next layer of the tree, which is more granular at the semantic level. We repeat this process until we obtain a tree structure $T$ with the root $r$, where semantically similar videos are located in the same path. Following previous work, we adopt cosine similarity as the metric for measuring the similarity between two video representations.

\noindent \textbf{Vi-SemTree for SemID Encoding.}
To directly retrieve a video set, we introduce SemID, a unique identifier derived from the Vi-SemTree, which serves as a pivotal reference for our generative model. We define the SemID as the path  $L_{\text{path}} = \{l_0, l_1, ..., l_d\}$, traversing from the root node $r$ down to a leaf node with $d$ marking tree depth. Specifically, the root node is represented as 0, serving as the start symbol. Each edge branching out from each node is numbered starting from 0, with values ranging from 0 to $k$. From top to bottom, this sequence serves as the SemID $L_{path}$.

\vspace{-5pt}

\subsection{Multi-view Textual Query Expansion}
Given video's rich semantic diversity, a single video can match multiple queries, but current models only generate limited descriptions, undermining semantic understanding. To address this, we introduced a multi-view query expansion strategy during Training stage (b) in Figure \ref{fig:model}, enhancing semantic depth.
Utilizing the Multimodal Large Language Model (MLLM), which excels in generating nuanced descriptions from various angles, we created 50 diverse queries per video, ensuring comprehensive semantic coverage. This approach surpasses traditional dense video caption models by offering detailed, multi-faceted insights into video content.


Based on the query $t$ provided in the dataset and the extended query $\hat{t}$, the query set $Q_i=\{t_i, \hat{t}_i\}$ is associated with each video $i$ in the training set $D$. We can represent the training data pairs for the generation model as$\{(q,\  SemID_i),  \ \  q \in Q_i,  \ \ i\in D \}$, where $SemID_i$ is the $L_{path} = \{l_0, l_1, ..., l_d\}$ of the node where video $i$ is located. However, as the tree structure deepens, the semantic of videos in different nodes becomes increasingly similar. This may lead to a decrease in the recall effect of relatively coarse-grained queries. To merge more videos with similar semantics and simplify the generation process, we choose the truncated version of the path as SemID, represented as $L_{path}^{trunc} = \{l_0, l_1, ..., l_{d-m}\}$, to ensure semantically consistent grouping. The $m$ means truncation length.  

\subsection{Generative Retrieval Model}
\begin{figure}[]
  \centering
  \includegraphics[width=\linewidth]{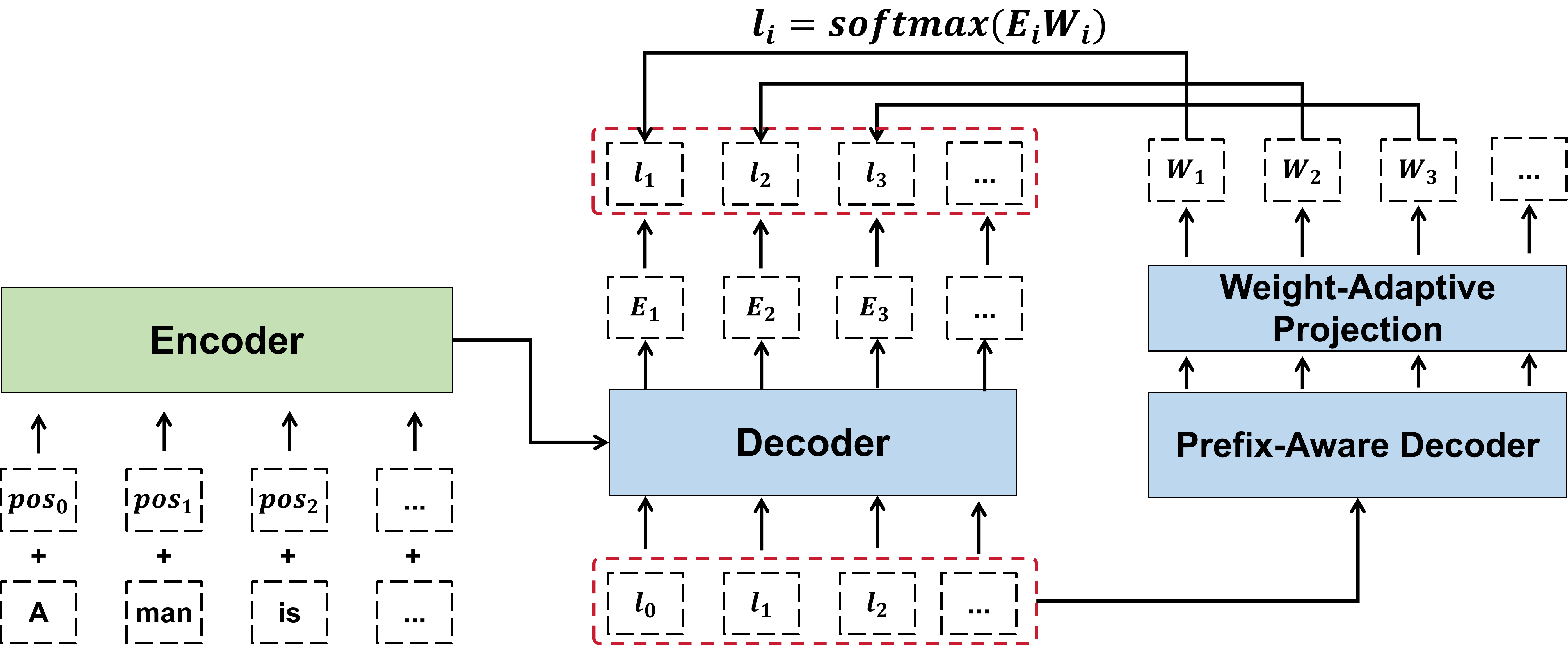}
  \caption{Overview of the generative model of T2VIndexer. Decoder uses different parameters when generating tokens at different positions.}
  \label{fig:gen}
  \vspace{-5pt}
\end{figure}
To achieve direct positioning of the target video through SemID, we chose to use a sequence-to-sequence generative model to directly generate the corresponding SemID based on the input query, as shown in Figure \ref{fig:gen}. First, the input query is added with position embedding and input into the transformer encoder to obtain the representation $f_t$. The probability of generating the SemID sequence as follows to construct, 
\begin{equation}
    p(L_{path}^{trunc}|f_t)=\prod_{i=1}p(l_i|f_t,l_0,l_1,...,l_{i-1})
    \label{eq:prob}
\end{equation}
which means the next token is generated according to the sequence of previously generated tokens. This probability problem can be solved by the traditional transformer encoder-decoder structure. The input of the encoder and the output of the decoder can be regarded as two different semantic spaces, corresponding to the natural language space and the video semantic tree space.

However, unlike standard decoding tasks, the same token appearing at different positions in the Vi-SemTree space has different meanings because they are in different layers of the Vi-SemTree. For example, as shown in the framework Figure \ref{fig:model}, for SemID `$0_0-0_1-0_2$', the token `$0_1$' in the first layer represents the semantic of the “sports”, while `$0_2$' in the second layer only represents “basketball”. In addition, even if they are in the same layer, the same token `$0_2$' in SemID `$0_0-0_1-0_2$' and `$0_0-2_1-0_2$' expresses different concept due to their different prefixes. To identify different representations at different positions during decoding, we were inspired by NCI \cite{DBLP:conf/nips/WangHWMWCXCZL0022} in document retrieval and used the Prefix-Aware Weight-Adaptor (PAWA) decoder, as shown in Figure \ref{fig:gen}.

Unlike the standard transformer decoder, the PAWA decoder uses different parameters when generating tokens at different positions, and the parameters between different steps are not shared. The specific settings are as follows. First, the encoder encodes the query to obtain the encoding $x$, and decoder output $E$ as follows, 
\begin{equation}
    E_i=\text{Decoder}(x,l_0,l_1,...,l_{i-1};\theta_i)
    \label{eq:dec}
\end{equation}

\noindent where $\theta_i$ represents the parameters of each decoding step, and the parameters of each step are different, distinguishing the semantic of different position tokens. In addition, to further enhance the prefix information as the basis for generation, the PAWA decoder further modifies the linear classification layers based on the prefix sequence. Specifically, instead of using the same projection weight $W$ in the linear classification layer, the PAWA decoder uses an additional decoder to generate different weights for each position,

\begin{equation}
    E_i' = \text{Decoder'}(l_0,l_1,...,l_{i-1};\theta_i')
\end{equation}

\begin{equation}
    W_i = \text{Linear}(E'_i)
\end{equation}

\noindent where $\theta_i'$ represents the parameters of the decoder that generates the weights, and $W_i$ is the generated weight matrix for the corresponding classifier. Finally, the i-th token is represented as $l_i$, calculated by $\text{softmax}(E_iW_i)$.

\subsection{Model Training and Inference}
\noindent \textbf{Training Loss.}
The loss function for a set of training examples $\mathcal{D}=\{(q, L_{path})\}$, consisting of queries (training queries and expansion queries) and video SemID, can be expressed as follows,
\begin{equation}
    \mathcal{L}(\theta)=\sum_{(q,L_{path})}^{\mathcal{D}}\text{log}(L_{path}|q,\theta)
\end{equation}
where $\text{log}(L_{path}|q,\theta)$ represents the probability of generating SemID based on $q$, and is a standard sequence-to-sequence cross-entropy loss with teacher forcing.

\noindent \textbf{Coarse-to-fine Inference.}
During the training stage, since the videos in the test set are not visible, this part of the videos is not assigned a SemID. In order to cover test set, a new video needs to be assigned a SemID as its identifier as shown in Figure \ref{fig:model} Inference Stage. The new video will get the representation in the same way as the training set using CLIP encoder, and then the video cosine similarity with the leaf nodes in the tree will be calculated. The video will be inserted into the leaf node with the highest similarity and inherit the SemID of the leaf node. For the input query $q$, the probability $p(L_{path}^{trunc}|q,\theta)$ is calculated by the generative model trained, and the target SemID is obtained to achieve direct positioning of a group of target videos. In the decoding process, considering the semantic summarization ability of natural language query, which makes videos that meet the description may not be in the same path, we use the Beam Search algorithm for decoding. This enables us to retrieve the top k SemIDs that meet the description, to adapt to this one-to-many issue. Based on the generated SemIDs, we obtain a small candidate set $V_{cand}=\{V_{SemID_1}, V_{SemID_2}, ..., V_{SemID_k}\}$ containing the target videos, where $SemID_i$ represents a group of videos corresponding to the i-th top SemID generated. 

Due to the limitations of the first stage, it is currently not feasible to achieve precise retrieval based on the query. In order to meet the requirements of the task and achieve precise retrieval, fine-tuning of $V_{cand}$ is required. For this purpose, we introduce a two-stage retrieval architecture and use the existing text-video retrieval model for precise retrieval in $V_{cand}$. T2VIndexer can be integrated with the existing retrieval model without adjusting the training parameters. For a text-video retrieval model $M_{base}$, the similarity $s$ between $q$ and each video $v$ in $V_{cand}$ is calculated, and the final retrieval result is obtained by ranking $s$. Overall, the entire retrieval process can be divided into two main stages: Pre-select based on generative models and Precise Recall based on contrastive learning models. 

\section{Experiments}
\label{sec:experiments}

\begin{table*}[]
\caption{Evaluation of the inference costs on MSR-VTT dataset. We report the R@1 metric for the text-to-video task. {\color[HTML]{000AFF} Blue} means stage 1 generative time cost, and {\color[HTML]{FE0000} red} means stage 2 time cost. The improvement of inference time refers to the compression ratio of T2VIndexer, which is obtained by dividing the inference time with T2VIndexer by the inference time of the baseline.}
\label{tab:eff}
\resizebox{\textwidth}{!}{%
\begin{tabular}{l|cc|cc|cc|cc}
\hline
                                  & \multicolumn{2}{c|}{\textbf{\#Candidate 1000}}    & \multicolumn{2}{c|}{\textbf{\#Candidate 3000}} & \multicolumn{2}{c|}{\textbf{\#Candidate 5000}} & \multicolumn{2}{c}{\textbf{\#Candidate 10000}} \\ \cline{2-9} 
\multirow{-2}{*}{\textbf{Method}} & \textbf{Inference Time(ms)↓}      & \textbf{R@1↑} & \textbf{Inference Time(ms)↓}  & \textbf{R@1↑}  & \textbf{Inference Time(ms)↓}  & \textbf{R@1↑}  & \textbf{Inference Time(ms)↓}  & \textbf{R@1↑}  \\ \hline
CLIP4Clip \cite{DBLP:journals/ijon/LuoJZCLDL22}                         & 220                               & 44.7          & 681                           & 36.2           & 1139                          & 31.6           & 2341                          & 25.4           \\
CLIP4Clip+T2VIndexer              & 105 ({\color[HTML]{000AFF} 52}+{\color[HTML]{FE0000} 53}) & 47.8          & 218 ({\color[HTML]{000AFF} 52}+{\color[HTML]{FE0000} 166})                   & 40.3           & 336 ({\color[HTML]{000AFF} 52}+{\color[HTML]{FE0000} 284})                   & 35.1           & 639 ({\color[HTML]{000AFF} 52}+{\color[HTML]{FE0000} 587})                   & 27.3           \\
Improvement                       & 0.47                              & 3.1           & 0.32                          & 4.1            & 0.29                          & 3.5            & 0.27                          & 1.9            \\ \hline
mPLUG \cite{DBLP:conf/icml/XuYYSYXLBQWXZH023}                            & 189                               & 53.0          & 562                           & 44.7           & 961                           & 38.3           & 1741                          & 29.1           \\
mPLUG+T2VIndexer                  & 97 ({\color[HTML]{000AFF} 52}+{\color[HTML]{FE0000} 45})                         & 54.3          & 189 ({\color[HTML]{000AFF} 52}+{\color[HTML]{FE0000} 137})                   & 45.8           & 290 ({\color[HTML]{000AFF} 52}+{\color[HTML]{FE0000} 238})                   & 40.4           & 485 ({\color[HTML]{000AFF} 52}+{\color[HTML]{FE0000} 433})                   & 30.1           \\
Improvement                       & 0.51                              & 1.3           & 0.34                          & 1.1            & 0.30                          & 2.1            & 0.28                          & 1.0            \\ \hline
CLIP-VIP \cite{xue2022clip}                          & 192                               & 54.1          & 579                           & 48.4           & 968                           & 40.1           & 1941                          & 31.5           \\
CLIP-VIP+T2VIndexer               & 98 ({\color[HTML]{000AFF} 52}+{\color[HTML]{FE0000} 46})                         & 55.1          & 194 ({\color[HTML]{000AFF} 52}+{\color[HTML]{FE0000} 142})                   & 49.3           & 295 ({\color[HTML]{000AFF} 52}+{\color[HTML]{FE0000} 243})                   & 41.3           & 537 ({\color[HTML]{000AFF} 52}+{\color[HTML]{FE0000} 485})                   & 33.1           \\
Improvement                       & 0.51                              & 1.0           & 0.34                          & 0.9            & 0.30                          & 1.2            & 0.28                         & 1.6            \\ \hline
\end{tabular}
}
\end{table*}

\begin{table*}[h]
\caption{Evaluation of the inference costs on MSVD, DiDeMo and ActivityNet dataset. We report the R@1 metric for the text-to-video task. {\color[HTML]{000AFF} Blue} means stage 1 generative time cost, and {\color[HTML]{FE0000} red} means stage 2 time cost.}
\label{tab:eff2}
\centering
\resizebox{0.8\textwidth}{!}{%
\begin{tabular}{l|cc|cc|cc}
\hline
\multirow{2}{*}{\textbf{Method}} & \multicolumn{2}{c|}{\textbf{MSVD}} & \multicolumn{2}{c|}{\textbf{DiDeMo}} & \multicolumn{2}{c}{\textbf{ActivityNet}} \\ \cline{2-7} 
                        & \textbf{Inference Time(ms)↓}           & \textbf{R@1↑}    & \textbf{Inference Time(ms)↓}            & \textbf{R@1↑}     & \textbf{Inference Time(ms)↓}              & \textbf{R@1↑}       \\ \hline
CLIP4Clip \cite{DBLP:journals/ijon/LuoJZCLDL22}              & 209             & 45.1    & 225              & 43.4     & 1041               & 40.2       \\
T2VIndexer+CLIP4Clip    & 106 ({\color[HTML]{000AFF} 52}+{\color[HTML]{FE0000} 54})     & 47.4    & 102 ({\color[HTML]{000AFF} 52}+{\color[HTML]{FE0000} 50})      & 46.3     & 311 ({\color[HTML]{000AFF} 52}+{\color[HTML]{FE0000} 259})       & 42.6       \\
Improvement             & 0.51            & 2.3     & 0.45             & 2.9      & 0.30               & 2.4        \\ \hline
mPLUG \cite{DBLP:conf/icml/XuYYSYXLBQWXZH023}                  & 193             & 53.6    & 201              & 56.4     & 923                & 52.2       \\
mPLUG+T2VIndexer        & 102 ({\color[HTML]{000AFF} 52}+{\color[HTML]{FE0000} 50})     & 55.4    & 98 ({\color[HTML]{000AFF} 52}+{\color[HTML]{FE0000} 46})       & 56.6     & 292 ({\color[HTML]{000AFF} 52}+{\color[HTML]{FE0000} 241})       & 53.5       \\
Improvement             & 0.53            & 1.8     & 0.48             & 0.2      & 0.32               & 1.3        \\ \hline
CLIP-VIP \cite{xue2022clip}               & 191             & 52.3    & 199              & 50.5     & 934                & 53.4       \\
CLIP-VIP+T2VIndexer     & 101 ({\color[HTML]{000AFF} 52}+{\color[HTML]{FE0000} 49})     & 54.0    & 97 ({\color[HTML]{000AFF} 52}+{\color[HTML]{FE0000} 45})       & 51.9     & 299 ({\color[HTML]{000AFF} 52}+{\color[HTML]{FE0000} 247})       & 54.9       \\
Improvement             & 0.53            & 1.7     & 0.49             & 1.4      & 0.32               & 1.5        \\ \hline
\end{tabular}
}
\end{table*}

\noindent \textbf{Datasets and Evaluation Metrics.}
We validate our model on four dataset: MSR-VTT, MSVD, DiDeMo, and ActivityNet Caption. MSR-VTT \cite{xu2016msr} encompasses 10,000 videos, paired with 200,000 captions. We employ the Training-9k variant, following the data splits proposed by \cite{DBLP:conf/eccv/Gabeur0AS20}. MSVD \cite{chen2011collecting} contains 1,970 videos, and a wealth of approximately 40 associated English sentences per video. Train, validation and test splits contain 1,200, 100, and 670 videos, respectively. DiDeMo \cite{anne2017localizing} contains 10,000 videos annotated with 40,000 sentences. We evaluate video-paragraph retrieval following \cite{DBLP:conf/bmvc/LiuANZ19}, \cite{lei2021less} and \cite{DBLP:conf/iccv/BainNVZ21}, where all sentence descriptions for a video are concatenated into a single query. ActivityNet \cite{DBLP:conf/iccv/KrishnaHRFN17} consists of 20,000 YouTube video. We follow the setting from \cite{zhang2018cross} to concatenate all the descriptions of a video to form a paragraph and evaluate the model with video-paragraph retrieval.

We adopt standard retrieval metrics, namely recall at rank K (R@K), calculates the percentage of instances where the correct result is successfully retrieved within top K.

\noindent \textbf{Implementation Details.}
We utilized the image encoder from the pre-trained CLIP (Vit B/32) model. For constructing the Vi-SemTree, we opted for $k$-means algorithm \cite{pedregosa2011scikit}, setting both k and c to 30. SemID truncation length $t$ set to 2 and select top 11 beam search results from generative model. The encoder parameters were initialized using the T5 pre-trained model \cite{DBLP:conf/nips/BrownMRSKDNSSAA20}, while the decoder parameters were randomly initialized. During training, the learning rate was set to $2 \times 10^{-4}$ for the encoder and $1 \times 10^{-4}$ for the decoder. We utilized 8 NVIDIA V100-32GB GPUs, with a batch size of 16 per GPU and a dropout ratio of 0.1. 

\vspace{-3pt}

\subsection{Efficiency of T2VIndexer}

The purpose of T2VIndexer is to improve retrieval efficiency while maintaining accuracy. In Table \ref{tab:eff}, we analyzed the accuracy and efficiency of T2VIndexer under different candidate sets with a single RTX3090 GPU and 8255C CPU. In the efficiency analysis, we imposed some restrictions to simulate real application scenarios. First, the time cost from receiving the query was calculated, without considering the offline phase, such as the construction of Vi-SemTree and the allocation of SemID. Second, each query in the Test set was retrieved one by one to return the target video, instead of obtaining all Queries and returning the overall results at once. From the Table \ref{tab:eff}, it can be seen that T2VIndexer significantly reduces inference time while maintaining the baseline effect. For example, compared with the traditional method under 1000 candidate videos, T2VIndexer reduces the time cost by 50\%. The efficiency improvement increases gradually with the size of the candidate set. Under 10,000 candidate videos, the time compression reaches 30\%. In addition, we also analyzed the performance of three other datasets, as shown in Table \ref{tab:eff2}.

Further analysis shows the efficiency improvement is due to the structure of Vi-SemTree. When the candidate set expands 10,000, the traditional method needs to process an additional 9000 data, perform a large number of similarity calculations and sorting. For T2VIndexer, the added 9000 videos will be distributed to each leaf node, and cost of generative model generating SemID remains unchanged, which ultimately improves efficiency. For example, if Vi-SemTree has 100 leaf nodes, for 10,000 candidate videos, there are an average of 100 target videos per leaf node, which significantly reduces the retrieval pressure of the baseline. This means that the T2VIndexer does not suffer from the same scalability issues as traditional methods, and allows for a more distribution of data.

\subsection{Evaluating on Large-Scale Dataset}
\begin{table}[]
\caption{Evaluation of the inference costs on large-scale dataset split from TGIF \cite{DBLP:conf/cvpr/LiSCTGJL16}. We report the R@1 metric for the text-to-video task. {\color[HTML]{000AFF} Blue} means stage 1 generative time cost, and {\color[HTML]{FE0000} red} means stage 2 time cost.}
\label{tab:largescale}
\begin{tabular}{l|cc}
\hline
\textbf{Method}      & \multicolumn{1}{l}{\textbf{Inference Time(ms)↓}} & \multicolumn{1}{l}{\textbf{R@1↑}} \\ \hline
Clip4Clip \cite{DBLP:journals/ijon/LuoJZCLDL22}           & 12322                                            & 13.4                              \\
Clip4Clip+T2VIndexer & 3064({\color[HTML]{000AFF} 52}+{\color[HTML]{FE0000} 3012})                                    & 16.5                              \\
Improvement          & 0.25                                             & 3.1                               \\ \hline
mPLUG \cite{DBLP:conf/icml/XuYYSYXLBQWXZH023}                & 10249                                            & 18.2                              \\
mPLUG+T2VIndexer     & 2831({\color[HTML]{000AFF} 52}+{\color[HTML]{FE0000} 2779})                                    & 21.1                              \\
Improvement          & 0.27                                             & 2.9                               \\ \hline
CLIP-VIP \cite{xue2022clip}            & 10414                                            & 19.7                              \\
CLIP-VIP+T2VIndexer  & 2902({\color[HTML]{000AFF} 52}+{\color[HTML]{FE0000} 2850})                                    & 22.4                              \\
Improvement          & 0.28                                             & 2.7                              \\ \hline
\end{tabular}

\end{table}
To further investigate the effectiveness of T2VIndexer in real retrieval scenarios, we evaluate on larger scale data. Due to the limited size of the existing dataset test sets, which mostly consist of 1000 candidate videos, we decided to redivide the TGIF dataset \cite{DBLP:conf/cvpr/LiSCTGJL16} into 50,000 training data and 50,000 testing data in a 5:5 ratio. Both the baseline and the T2VIndexer generative model will be trained solely on the 50,000 training data. Table \ref{tab:largescale} displays our test results, with each block representing a set of test results. It is evident from the results that in large-scale retrieval scenarios, T2VIndexer demonstrates more significant improvements in both efficiency and accuracy compared to smaller-scale data.

The improvements in both efficiency and accuracy primarily stem from the pre-select mechanism employed by T2VIndexer. It generates an ID sequence with constant-time complexity, targeting a small subset of videos. The base model then performs similarity calculations and sorting only within this subset, significantly cutting computational load by bypassing individual matching and boosting retrieval speed. Additionally, the pre-select mechanism accurately identifies the target video, preemptively discarding most irrelevant videos and reducing noise, thus acting as a filter that enhances baseline accuracy.
\vspace{-3pt}
\subsection{State-of-the-Art Comparison}
\begin{table*}[]
\caption{Comparison with Existing One-stream approaches and Two-stream approaches. Our Re-implemented methods are denoted by the superscript ‘*’. The highest retrieval recall in each block is marked with \underline{underline}. The recall of our models is marked with blue color when it is better than the baseline model.}
\label{tab:sota}
\resizebox{\textwidth}{!}{%
\begin{tabular}{lcccccccccccccccc}
\hline
\multicolumn{1}{l|}{}                                              & \multicolumn{4}{c|}{\textbf{MSR-VTT 1k}}                                                                                                                                                     & \multicolumn{4}{c|}{\textbf{MSVD}}                                                                                                                                                           & \multicolumn{4}{c|}{\textbf{DiDeMo}}                                                                                                                                                         & \multicolumn{4}{c}{\textbf{ActivityNet Caption}}                                                                                                \\ \cline{2-17} 
\multicolumn{1}{l|}{\multirow{-2}{*}{\textbf{Methods}}}            & \multicolumn{1}{c|}{\textbf{R@1}} & \multicolumn{1}{c|}{\textbf{R@5}} & \multicolumn{1}{c|}{\textbf{R@10}} & \multicolumn{1}{c|}{\textbf{R@sum}}                                             & \multicolumn{1}{c|}{\textbf{R@1}} & \multicolumn{1}{c|}{\textbf{R@5}} & \multicolumn{1}{c|}{\textbf{R@10}} & \multicolumn{1}{c|}{\textbf{R@sum}}                                             & \multicolumn{1}{c|}{\textbf{R@1}} & \multicolumn{1}{c|}{\textbf{R@5}} & \multicolumn{1}{c|}{\textbf{R@10}} & \multicolumn{1}{c|}{\textbf{R@sum}}                                             & \multicolumn{1}{c|}{\textbf{R@1}} & \multicolumn{1}{c|}{\textbf{R@5}} & \multicolumn{1}{c|}{\textbf{R@10}} & \textbf{R@sum}                     \\ \hline
\multicolumn{17}{c}{\textbf{One-stream approaches}}                                                                                                                                                                                                                                                                                                                                                                                                                                                                                                                                                                                                                                                                                                                                                                 \\ \hline
\multicolumn{1}{l|}{UniVL \cite{luo2020univl}}                                         & 21.2                              & 49.6                              & 63.1                               & \multicolumn{1}{c|}{133.9}                                                      & -                                 & -                                 & -                                  & \multicolumn{1}{c|}{-}                                                          & -                                 & -                                 & -                                  & \multicolumn{1}{c|}{-}                                                          & -                                 & -                                 & -                                  & -                                  \\
\multicolumn{1}{l|}{ClipBERT \cite{lei2021less}}                                      & 22.0                              & 46.8                              & 59.9                               & \multicolumn{1}{c|}{128.7}                                                      & -                                 & -                                 & -                                  & \multicolumn{1}{c|}{-}                                                          & {\ul 20.4}                        & {\ul 48.0}                        & {\ul 60.8}                         & \multicolumn{1}{c|}{{\ul 129.2}}                                                & {\ul 21.3}                        & {\ul 49.0}                        & {\ul 63.5}                         & {\ul 133.8}                        \\
\multicolumn{1}{l|}{VLM \cite{xu2021vlm}}                                           & {\ul 28.1}                        & {\ul 55.5}                        & {\ul 67.4}                         & \multicolumn{1}{c|}{{\ul 151.0}}                                                & -                                 & -                                 & -                                  & \multicolumn{1}{c|}{-}                                                          & -                                 & -                                 & -                                  & \multicolumn{1}{c|}{-}                                                          & -                                 & -                                 & -                                  & -                                  \\ \hline
\multicolumn{17}{c}{\textbf{Two-stream approaches}}                                                                                                                                                                                                                                                                                                                                                                                                                                                                                                                                                                                                                                                                                                                                                             \\ \hline
\multicolumn{1}{l|}{MMT \cite{DBLP:conf/eccv/Gabeur0AS20}}                                           & 26.6                              & 57.1                              & 69.6                               & \multicolumn{1}{c|}{153.3}                                                      & -                                 & -                                 & -                                  & \multicolumn{1}{c|}{-}                                                          & -                                 & -                                 & -                                  & \multicolumn{1}{c|}{-}                                                          & -                                 & 61.4                              & -                                  & -                                  \\
\multicolumn{1}{l|}{Frozen \cite{DBLP:conf/iccv/BainNVZ21}}                                        & 31.0                              & 59.5                              & 70.5                               & \multicolumn{1}{c|}{161.0}                                                      & 33.7                              & 64.7                              & 76.3                               & \multicolumn{1}{c|}{174.7}                                                      & 34.6                              & 65.0                              & 74.7                               & \multicolumn{1}{c|}{174.3}                                                      & 28.8                              & 60.9                              & -                                  & -                                  \\
\multicolumn{1}{l|}{CLIP4Clip \cite{DBLP:journals/ijon/LuoJZCLDL22}}                                     & 44.5                              & 71.4                              & 81.6                               & \multicolumn{1}{c|}{197.5}                                                      & 45.2                              & 75.5                              & 84.3                               & \multicolumn{1}{c|}{205.0}                                                      & 43.4                              & 70.2                              & 80.6                               & \multicolumn{1}{c|}{194.2}                                                      & 40.5                              & 72.4                              & -                                  & -                                  \\
\multicolumn{1}{l|}{CAMoE \cite{DBLP:journals/corr/abs-2109-04290}}                                         & 44.6                              & 72.6                              & 81.8                               & \multicolumn{1}{c|}{199.0}                                                      & 46.9                              & 76.1                              & 85.5                               & \multicolumn{1}{c|}{208.5}                                                      & -                                 & -                                 & -                                  & \multicolumn{1}{c|}{-}                                                          & -                                 & -                                 & -                                  & -                                  \\
\multicolumn{1}{l|}{CLIP2Video \cite{DBLP:journals/corr/abs-2106-11097}}                                    & 45.6                              & 72.6                              & 81.7                               & \multicolumn{1}{c|}{199.9}                                                      & 47.0                              & 76.8                              & 85.9                               & \multicolumn{1}{c|}{209.7}                                                      & -                                 & -                                 & -                                  & \multicolumn{1}{c|}{-}                                                          & -                                 & -                                 & -                                  & -                                  \\
\multicolumn{1}{l|}{Cap4Video \cite{DBLP:conf/cvpr/WuLFWO23}}                                     & 51.4                              & 75.7                              & 83.9                               & \multicolumn{1}{c|}{211.0}                                                      & {\ul 51.8}                        & {\ul 80.8}                        & {\ul 88.3}                         & \multicolumn{1}{c|}{{\ul 220.9}}                                                & 52.0                              & 79.4                              & 87.5                               & \multicolumn{1}{c|}{218.9}                                                      & -                                 & -                                 & -                                  & -                                  \\
\multicolumn{1}{l|}{mPLUG \cite{DBLP:conf/icml/XuYYSYXLBQWXZH023}}                                         & 53.1                              & {\ul 77.6}                        & 84.7                               & \multicolumn{1}{c|}{215.4}                                                      & -                                 & -                                 & -                                  & \multicolumn{1}{c|}{-}                                                          & {\ul 56.4}                        & {\ul 79.1}                        & 85.2                               & \multicolumn{1}{c|}{{\ul 220.7}}                                                & -                                 & -                                 & -                                  & -                                  \\
\multicolumn{1}{l|}{CLIP-VIP \cite{xue2022clip}}                                      & {\ul 54.2}                        & 77.2                              & {\ul 84.8}                         & \multicolumn{1}{c|}{{\ul 216.2}}                                                & -                                 & -                                 & -                                  & \multicolumn{1}{c|}{-}                                                          & 50.5                              & 78.4                              & {\ul 87.1}                         & \multicolumn{1}{c|}{216.0}                                                      & {\ul 53.4}                        & {\ul 81.4}                        & {\ul 90.0}                         & {\ul 224.8}                        \\ \hline
\multicolumn{17}{c}{\textbf{Ours}}                                                                                                                                                                                                                                                                                                                                                                                                                                                                                                                                                                                                                                                                                                                                                                                \\ \hline
\multicolumn{1}{l|}{CLIP4Clip*}                                    & 44.5                              & 71.0                              & 81.6                               & \multicolumn{1}{c|}{197.1}                                                      & 45.1                              & 75.6                              & 83.9                               & \multicolumn{1}{c|}{204.6}                                                      & 43.4                              & 70.1                              & 80.1                               & \multicolumn{1}{c|}{193.6}                                                      & 40.2                              & 72.4                              & 80.4                               & 193.0                              \\
\rowcolor[HTML]{CFCFCF} 
\multicolumn{1}{l|}{\cellcolor[HTML]{CFCFCF}T2VIndexer+CLIP4Clip*} & {\color[HTML]{000AFF} 47.8}       & {\color[HTML]{000AFF} 72.2}       & {\color[HTML]{000AFF} 82.4}        & \multicolumn{1}{c|}{\cellcolor[HTML]{CFCFCF}{\color[HTML]{000AFF} 202.4}}       & {\color[HTML]{000AFF} 47.4}       & {\color[HTML]{000AFF} 76.4}       & {\color[HTML]{000AFF} 85.1}        & \multicolumn{1}{c|}{\cellcolor[HTML]{CFCFCF}{\color[HTML]{000AFF} 208.9}}       & {\color[HTML]{000AFF} 46.3}       & {\color[HTML]{000AFF} 72.4}       & {\color[HTML]{000AFF} 83.1}        & \multicolumn{1}{c|}{\cellcolor[HTML]{CFCFCF}{\color[HTML]{000AFF} 199.8}}       & {\color[HTML]{000AFF} 42.6}       & {\color[HTML]{000AFF} 73.5}       & {\color[HTML]{000AFF} 80.9}        & {\color[HTML]{000AFF} 197.0}       \\
\multicolumn{1}{l|}{Improvement}                                   & \textbf{+3.3}                     & \textbf{+2.2}                     & \textbf{+0.8}                      & \multicolumn{1}{c|}{\textbf{+5.3}}                                              & \textbf{+2.3}                     & \textbf{+0.8}                     & \textbf{+1.2}                      & \multicolumn{1}{c|}{\textbf{+4.3}}                                              & \textbf{+2.9}                     & \textbf{+2.3}                     & \textbf{+1.0}                      & \multicolumn{1}{c|}{\textbf{+6.2}}                                              & \textbf{+2.4}                     & \textbf{+1.1}                     & \textbf{+0.5}                      & \textbf{+4.0}                      \\ \hline
\multicolumn{1}{l|}{mPLUG*}                                        & 53.0                              & 77.4                              & 82.3                               & \multicolumn{1}{c|}{212.7}                                                      & 53.6                              & 81.4                              & 88.3                               & \multicolumn{1}{c|}{223.3}                                                      & 56.4                              & 79.0                              & 84.3                               & \multicolumn{1}{c|}{219.7}                                                      & 52.2                              & 80.8                              & 89.3                               & 222.3                              \\
\rowcolor[HTML]{CFCFCF} 
\multicolumn{1}{l|}{\cellcolor[HTML]{CFCFCF}T2VIndexer+mPLUG*}     & {\color[HTML]{000AFF} 54.3}       & {\color[HTML]{000AFF} {\ul 77.7}} & {\color[HTML]{000AFF} 82.5}        & \multicolumn{1}{c|}{\cellcolor[HTML]{CFCFCF}{\color[HTML]{000AFF} 214.5}}       & {\color[HTML]{000AFF} {\ul 55.4}} & {\color[HTML]{000AFF} {\ul 81.9}} & {\color[HTML]{000AFF} {\ul 88.5}}  & \multicolumn{1}{c|}{\cellcolor[HTML]{CFCFCF}{\color[HTML]{000AFF} {\ul 225.8}}} & {\color[HTML]{000AFF} {\ul 56.6}} & {\color[HTML]{000AFF} 79.1}       & {\color[HTML]{333333} 84.1}        & \multicolumn{1}{c|}{\cellcolor[HTML]{CFCFCF}{\color[HTML]{000AFF} {\ul 219.8}}} & {\color[HTML]{000AFF} 53.5}       & {\color[HTML]{000AFF} 81.1}       & {\color[HTML]{000AFF} 89.5}        & {\color[HTML]{000AFF} 224.1}       \\
\multicolumn{1}{l|}{Improvement}                                   & \textbf{+1.3}                     & \textbf{+0.3}                     & \textbf{+0.2}                      & \multicolumn{1}{c|}{\textbf{+1.8}}                                              & \textbf{+1.8}                     & \textbf{+0.5}                     & \textbf{+0.2}                      & \multicolumn{1}{c|}{\textbf{+2.5}}                                              & \textbf{+0.2}                     & \textbf{+0.1}                     & \textbf{-0.2}                      & \multicolumn{1}{c|}{\textbf{+0.1}}                                              & \textbf{+1.3}                     & \textbf{+0.3}                     & \textbf{+0.2}                      & \textbf{+1.8}                      \\ \hline
\multicolumn{1}{l|}{CLIP-VIP*}                                     & 54.1                              & 77.0                              & 84.7                               & \multicolumn{1}{c|}{215.8}                                                      & 52.3                              & 81.6                              & 88.2                               & \multicolumn{1}{c|}{222.1}                                                      & 50.5                              & 78.3                              & 86.6                               & \multicolumn{1}{c|}{215.4}                                                      & 53.4                              & 82.3                              & 89.7                               & 225.4                              \\
\rowcolor[HTML]{CFCFCF} 
\multicolumn{1}{l|}{\cellcolor[HTML]{CFCFCF}T2VIndexer+CLIP-VIP*}  & {\color[HTML]{000AFF} {\ul 55.1}} & {\color[HTML]{000AFF} 77.2}       & {\color[HTML]{000AFF} {\ul 85.0}}  & \multicolumn{1}{c|}{\cellcolor[HTML]{CFCFCF}{\color[HTML]{000AFF} {\ul 217.3}}} & {\color[HTML]{000AFF} 54.0}       & {\color[HTML]{333333} 81.3}       & {\color[HTML]{000AFF} 88.3}        & \multicolumn{1}{c|}{\cellcolor[HTML]{CFCFCF}{\color[HTML]{000AFF} 223.6}}       & {\color[HTML]{000AFF} 51.9}       & {\color[HTML]{000AFF} {\ul 79.2}} & {\color[HTML]{000AFF} {\ul 87.1}}  & \multicolumn{1}{c|}{\cellcolor[HTML]{CFCFCF}{\color[HTML]{000AFF} 218.2}}       & {\color[HTML]{000AFF} {\ul 54.9}} & {\color[HTML]{000AFF} {\ul 82.5}} & {\color[HTML]{000AFF} {\ul 90.0}}  & {\color[HTML]{000AFF} {\ul 227.4}} \\
\multicolumn{1}{l|}{Improvement}                                   & \textbf{+1.0}                     & \textbf{+0.2}                     & \textbf{+0.3}                      & \multicolumn{1}{c|}{\textbf{+1.5}}                                              & \textbf{+1.7}                     & \textbf{-0.3}                     & \textbf{+0.1}                      & \multicolumn{1}{c|}{\textbf{+1.5}}                                              & \textbf{+1.4}                     & \textbf{+0.9}                     & \textbf{+0.5}                      & \multicolumn{1}{c|}{\textbf{+2.8}}                                              & \textbf{+1.5}                     & \textbf{+0.2}                     & \textbf{+0.3}                      & \textbf{+2.0}                      \\ \hline
\end{tabular}
}
\end{table*}

Table \ref{tab:sota} is divided into three sections: Interactive, Independent, and our T2VIndexer, each based on various baselines. Two-stream methods in the first two sections typically perform well due to CLIP's strong pretraining. In the third section, T2VIndexer was built using CLIP4CLIP, mPLUG, and CLIP-VIP. These models' performance improved sequentially, allowing T2VIndexer to demonstrate varying effectiveness. On the MSR-VTT dataset, T2VIndexer boosted CLIP4CLIP's R@1 by 3.3\%, but its accuracy gain lessened with better baselines, peaking at a 1.0\% increase with CLIP-VIP.


This can be attributed to the candidates provided by T2VIndexer to baseline models. T2VIndexer provides the same set of candidates to the baseline models for a given query, effectively eliminating a considerable number of irrelevant videos. For lower-performing models, this removal of irrelevant videos is useful, significantly reducing the input noise and enhancing accuracy. However, higher-performing models are less sensitive to noise and can distinguish relevant content more effectively, resulting in relatively lower improvements when assisted by T2VIndexer.

\subsection{Ablation Study on Model Structure}

\begin{table}[]
\caption{Ablation Study on MSR-VTT-1kA}
\vspace{-5pt}
\label{tab:abl}
\resizebox{\linewidth}{!}{%
\begin{tabular}{l|ccc}
\hline
\textbf{Method}                           & \multicolumn{1}{l}{\textbf{R@1}} & \multicolumn{1}{l}{\textbf{R@5}} & \multicolumn{1}{l}{\textbf{R@10}} \\ \hline
\textbf{T2VIndexer+CLIP-VIP (full model)} & \textbf{55.1}                         & \textbf{77.2}                         & \textbf{85.0}                          \\
w/o Query expansion                       & 45.2                                  & 69.3                                  & 77.9                                   \\
w/o Multi-view query expansion            & 52.8                                  & 75.3                                  & 83.1                                   \\
w/o Vi-SemTree and SemID                  & 54.0                                  & 76.9                                  & 84.6                                  \\ \hline
\end{tabular}
}
\vspace{-5pt}
\end{table}

To further investigate the impact of different components on the model's performance, we report the ablation results on the MSR-VTT dataset in Table \ref{tab:abl}. (1) \textbf{Without query expansion (w/o query expansion)} This means that only the captions provided in the training set are used as the training basis. This part has the most significant impact on the model's results. During the SemID generation process by T2VIndexer, the original video is not seen. If the semantics of the original video are not injected into SemID through queries during the training phase, the model will fail to establish a relationship between the text and SemID and will not be able to correctly generate SemID for queries not seen during training. (2) \textbf{Without Multi-view query expansion (w/o Multi-view query expansion)} This indicates not using an MLLM (Multilingual Language Models) to generate multi-view descriptions, and only using a dense caption model for generating descriptions. The results suggest that the semantic expansion of SemID allows the model to learn richer information, which can better apply SemID to the test set. (3) \textbf{Without Vi-SemTree and SemID (w/o Vi-SemTree and SemID)} This means that a single $K$-means clustering is performed, and each cluster is assigned a number as the identifier for the videos. The experiment confirms our theoretical premise that structured pre-injection of prior knowledge facilitates superior generalization.

\subsection{Ablation Study on Different MLLMs}
\begin{table}[t]
\caption{Ablation Study on MLLMs}
\label{tab:ablm}
\vspace{-5pt}
\centering
\begin{tabular}{l|ccc}
\hline
MLLM      & R@1  & R@5  & R@10 \\ \hline
mPLUG-owl   & 55.1 & 77.2 & 85.0 \\  
Minigpt-4  & 55.1 & 77.0 & 85.1 \\
LLaVA     & 55.3 & 77.5 & 85.3 \\ \hline
\end{tabular}
\vspace{-5pt}
\end{table}
Utilizing Multi-Modal Large Language Models (MLLMs) for query expansion effectively enhances the model's generalization capabilities. Various MLLMs show little difference in the quality of query expansion, so the model's effectiveness does not rely on a specific MLLM. Apart from mPLUG-owl tested in the paper, we have also conducted supplementary tests with Minigpt-4 and LLaVA. As evident from the results in Table \ref{tab:ablm} on the MSR-VTT dataset, different MLLMs have a minor impact on retrieval accuracy.

\subsection{Ablation Study on Video Feature Extractor}
\begin{table}[]
\caption{Ablation Study on different video feature extractors on MSR-VTT with CLIP-VIP as baseline.}
\label{tab:ext}
\centering
\begin{tabular}{l|ccc}
\hline
Feature Extractor  & R@1  & R@5  & R@10 \\ \hline
S3D                & 51.3 & 72.4 & 81.6 \\
CLIP image encoder & 55.1 & 77.2 & 85.0 \\
VideoMAE           & 55.4 & 77.8 & 85.6 \\ \hline
\end{tabular}
\vspace{-5pt}
\end{table}
Different methods of video feature extraction will affect the structure of the Vi-Sem tree, thereby influencing the overall model's effectiveness. In Table\ref{tab:ext}, we tested three different video feature extraction methods: S3D based on pixel features, CLIP image encoder based on semantic features of frames, and VideoMAE pre-trained for video recognition. From the results comparison, it can be observed that the method based on pixel features performs the worst, even falling below the baseline model CLIP-VIP. This is due to its tree structure reflecting pixel info rather than relevant semantic info, leading to poor performance in the video pre-select stage and inability to return correct video clusters for precise recall.

\subsection{Generative Result Visualization}
\begin{figure}[h]
  \centering
  \includegraphics[width=\linewidth]{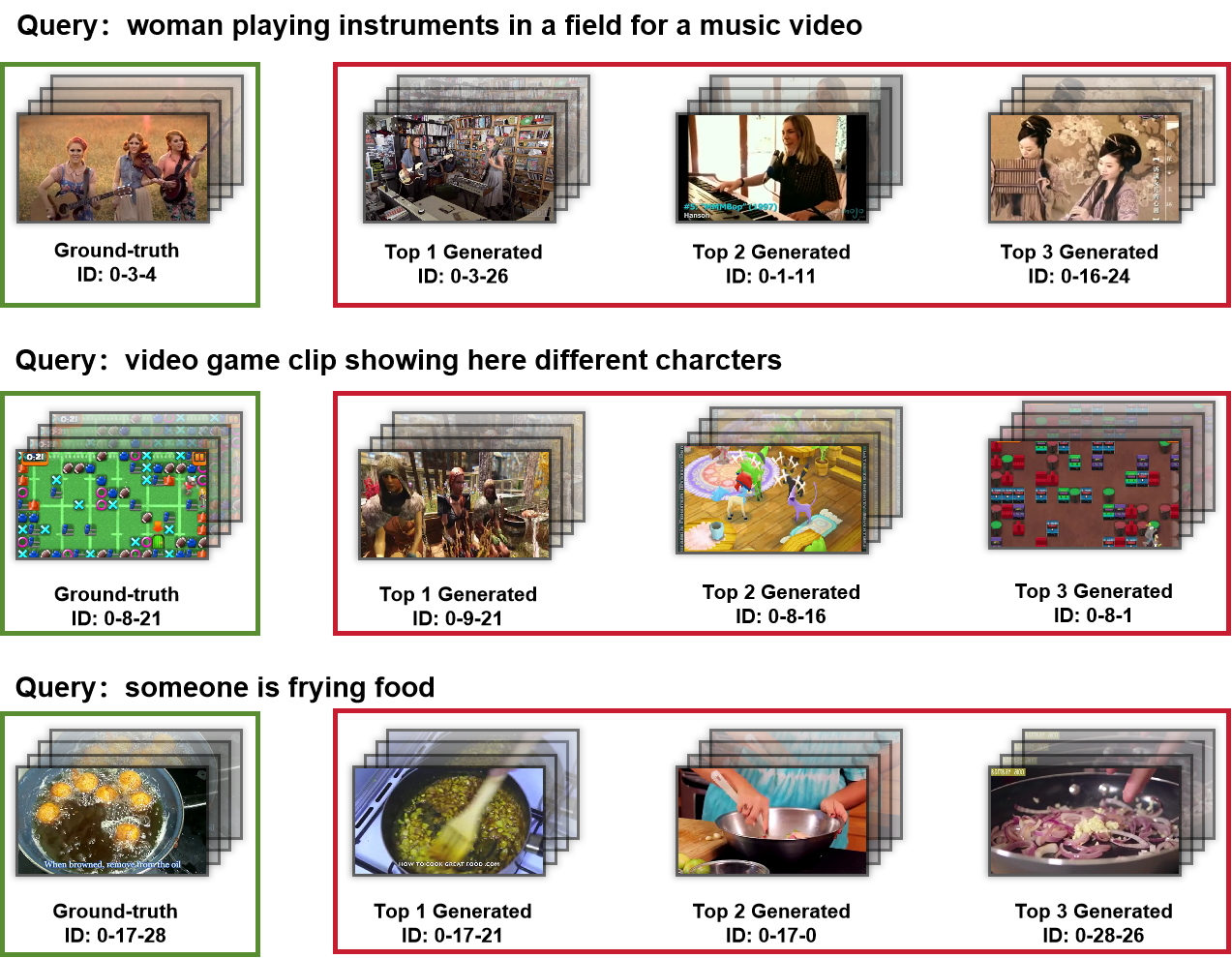}
  \caption{Visualization of the difference between Generative Results and Ground-truth. We show the top-3 Generated SemIDs for each text query. The truly matched results are marked in green boxes and the falsely matched results are in red boxes.}
  \label{fig:vis}
  \vspace{-5pt}
\end{figure}

We further demonstrated the ability of T2VIndexer to locate target videos through visualization. Three examples are shown in Figure \ref{fig:vis}, where the SemID generated by T2VIndexer has a high semantic similarity with the target SemID, even in wrong mapping cases. Specifically, for the query “video game clip showing here different characters”, the SemID generated by T2VIndexer, 0-9-21, has a stronger matching relationship with the query than ground-truth video. This indicates that the model has effectively learned the mapping between natural language space and SemID space, achieving retrieve target videos directly.

\subsection{Parameter Analysis}
\begin{figure}[h]
\vspace{-5pt}
  \centering
  \includegraphics[width=\linewidth]{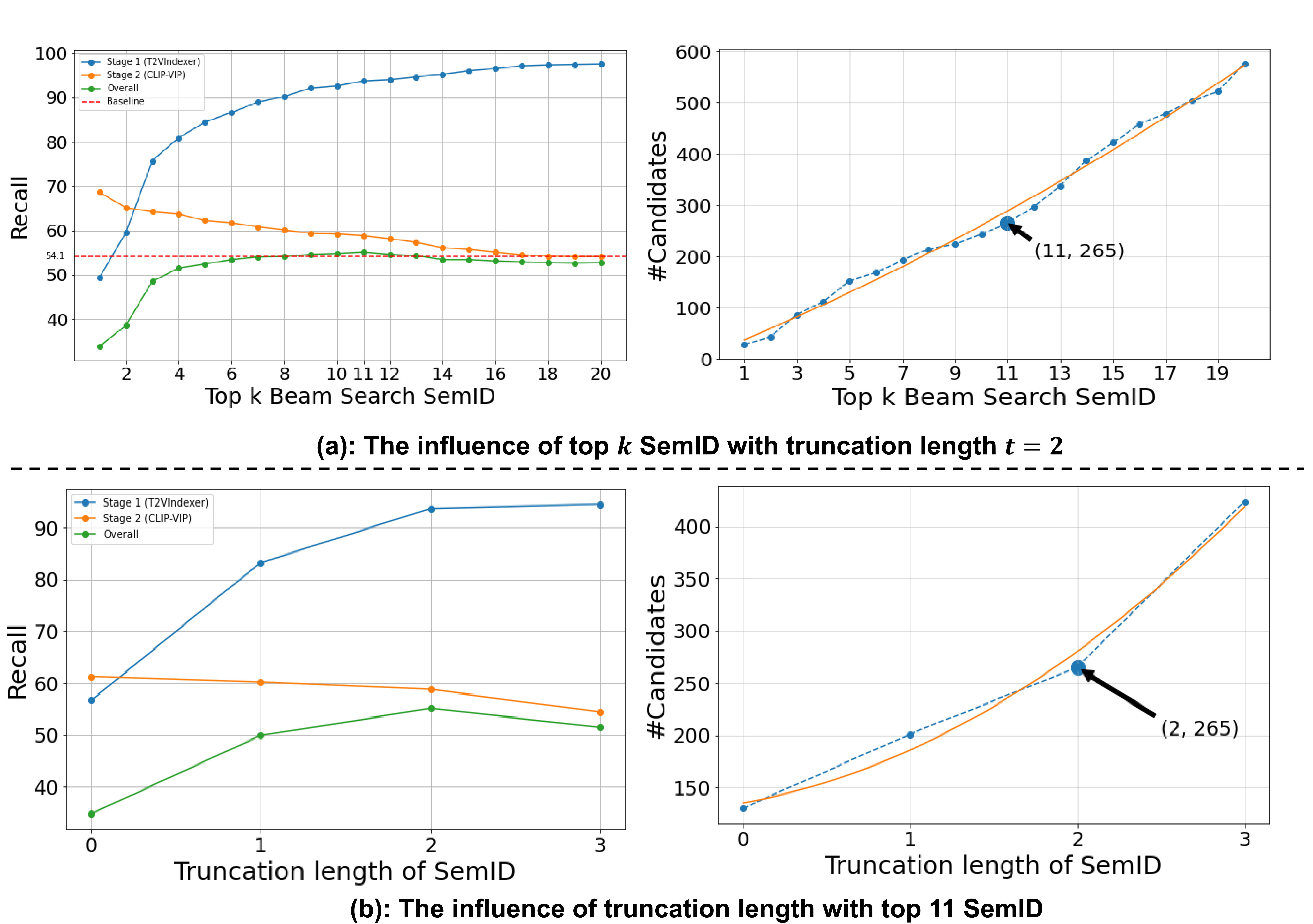}
  \caption{Different truncation length and top k SemID for T2VIndexer on MSR-VTT-1kA.}
  \label{fig:para}
\end{figure}
We analyzed various model setups. Figure \ref{fig:para} (a) illustrates the top $k$ SemIDs selected via beam search at $t=2$, affects recall and candidate count. As $k$ grows, second-stage recall dips due to more candidates, but overall recall peaks at 55.1\% with $k=11$. Figure \ref{fig:para} (b) examines the impact of different $t$ values on recall and candidate sets when $k=11$, showing optimal results at $t=2$.

\section{Limitation and Future Work}
\begin{table}[]
\caption{Ensemble framework comparative analysis. Pre-select size represents the number of videos pre-selected in the first stage.}
\label{tab:lim}
\resizebox{\linewidth}{!}{%
\begin{tabular}{l|ccc|ccc}
\hline
\multirow{2}{*}{Ensemble framework} & \multicolumn{3}{c|}{Pre-select size 50}                                                                                                                                          & \multicolumn{3}{c}{Pre-select size 265}                                                                                                                                          \\ \cline{2-7} 
                                    & \begin{tabular}[c]{@{}c@{}}Stage 1\\ Recall\end{tabular} & \begin{tabular}[c]{@{}c@{}}Overall\\ R@1\end{tabular} & \begin{tabular}[c]{@{}c@{}}Inference\\ Time (ms)\end{tabular} & \begin{tabular}[c]{@{}c@{}}Stage 1\\ Recall\end{tabular} & \begin{tabular}[c]{@{}c@{}}Overall\\ R@1\end{tabular} & \begin{tabular}[c]{@{}c@{}}Inference\\ Time (ms)\end{tabular} \\ \hline
mPLUG+CLIP-VIP                      & \textbf{93.4}                                            & \textbf{55.6}                                         & 198(189+9)                                                    & \textbf{97.2}                                            & \textbf{57.8}                                         & 235(189+46)                                                   \\
T2VIndexer+CLIP-VIP                 & 42.9                                                     & 21.4                                                  & \textbf{61(52+9)}                                             & 93.7                                                     & 55.1                                                  & \textbf{98(52+46)}                                            \\ \hline
\end{tabular}
}
\vspace{-15pt}
\end{table}
Although T2VIndexer has achieved certain results in efficient text-video retrieval, there still exists performance limitations. It uses a two-stage process: pre-select and precise retrieval, akin to an ensemble method. We compared it with a two-stage ensemble of mPLUG and CLIP-VIP, shown in Table \ref{tab:lim}. While ensemble methods enhance accuracy at the cost of efficiency, they outperform T2VIndexer's generative approach. However, T2VIndexer excels in efficiency by directly targeting candidate sets, unlike existing models that process all candidates. Additionally, existing pipelines struggle with new videos, requiring costly similarity calculations and sorting for each insertion into Vi-SemTree, averaging 200 ms per video in the MSR-VTT test set. Our future efforts will enhance the generative stage's accuracy for better precision and aim to cut new data processing time and boost flexibility.

\section{Conclusion}
In this paper, we propose T2VIndexer, a model-based video indexer that generates video identifiers directly and retrieves candidate videos with constant time complexity, in order to shorten the overall retrieval time while maintaining the retrieval accuracy of the base model. We use hierarchical clustering to organize videos into a tree structure called Vi-SemTree, which contains multiple layers corresponding to relationships from coarse to fine. We specifically trained a generative model for Vi-SemTree paths, correctly mapping natural language space and video semantic space. T2VIndexer is model-independent and can be seamlessly integrated with existing methods. However, the retrieval effect of the generative model is currently limited, and our future work will focus on improving the accuracy of the generative retrieval to achieve precise retrieval, and further improve the flexibility of the model when receiving new videos and reduce preprocessing time.
\begin{acks}
This work was supported by the Central Guidance for Local Special Project (Grant No. Z231100005923044) and the Climbing Plan Project (Grant No. E3Z0261).
\end{acks}

\bibliographystyle{ACM-Reference-Format}
\balance
\bibliography{sample-base}










\end{document}